\documentclass[10pt,twocolumn,letterpaper]{article}

\usepackage{iccv}

\usepackage{times}
\usepackage{epsfig}
\usepackage{graphicx}
\usepackage{epstopdf}
\usepackage{amsmath}
\usepackage{amssymb}
\usepackage{multirow}
\usepackage{bbding}
\usepackage{commath}
\usepackage{subcaption}
\usepackage{gensymb}

\usepackage{xcolor}
\usepackage{bbding}
\usepackage{pifont}

\DeclareMathOperator*{\argmaxB}{argmax}

\usepackage[pagebackref=true,breaklinks=true,letterpaper=true,colorlinks,bookmarks=false]{hyperref}

\iccvfinalcopy 

\def\iccvPaperID{2037} 

\ificcvfinal\fi
\begin{document}

\title{Fashion Retrieval via Graph Reasoning Networks on a Similarity Pyramid}

\author{
Zhanghui Kuang$^{1}$\thanks{They contributed equally to this work},\hspace{1px} Yiming Gao$^{12*}$,\hspace{1px} Guanbin Li$^2$,\hspace{1px} Ping Luo$^3$,\hspace{1px} Yimin Chen$^1$,\hspace{1px} Liang Lin$^2$,\hspace{1px} Wayne Zhang$^1$\thanks{Wayne Zhang is the corresponding author}\\
$^1$SenseTime Research \hspace{3px} $^2$Sun Yat-sen University \hspace{3px} $ ^3$The University of Hong Kong\\
\tt\small \{kuangzhanghui, chenyimin,  wayne.zhang\}@sensetime.com \hspace{2px} gaoym9@mail2.sysu.edu.cn \hspace{2px} \\
\tt\small liguanbin@mail.sysu.edu.cn \hspace{2px} pluo@cs.hku.hk \hspace{2px} linliang@ieee.org
}
\maketitle
\thispagestyle{empty}

\begin{abstract}
 Matching clothing images from customers and online shopping stores has rich applications in E-commerce. Existing algorithms encoded an image as a global feature vector and performed retrieval with the global representation. However, discriminative local information on clothes are submerged in this global representation, resulting in sub-optimal performance.
 To address this issue, we propose a novel Graph Reasoning Network (GRNet) on a Similarity Pyramid, which learns similarities between a query and a gallery cloth by using both global and local representations in multiple scales.
  The similarity pyramid is represented by a Graph of similarity, where nodes represent similarities between clothing components at different scales, and the final matching score is obtained by message passing along edges. In GRNet, graph reasoning is solved by training a graph convolutional network, enabling to align salient clothing components to improve clothing retrieval.
  To facilitate future researches,
  we introduce a new benchmark FindFashion, containing rich annotations of bounding boxes, views, occlusions, and cropping.
  Extensive experiments show that GRNet obtains new state-of-the-art results on two challenging benchmarks, \eg pushing the top-1, top-20, and top-50 accuracies on DeepFashion to 26\%, 64\%, and 75\% (\ie 4\%, 10\%, and 10\% absolute improvements), outperforming competitors with large margins. On FindFashion, GRNet achieves considerable improvements on all empirical settings.
 \end{abstract}

\section{Introduction}
\begin{figure}
\begin{minipage}[t]{1\linewidth}
\centering
\includegraphics[width =1\textwidth]{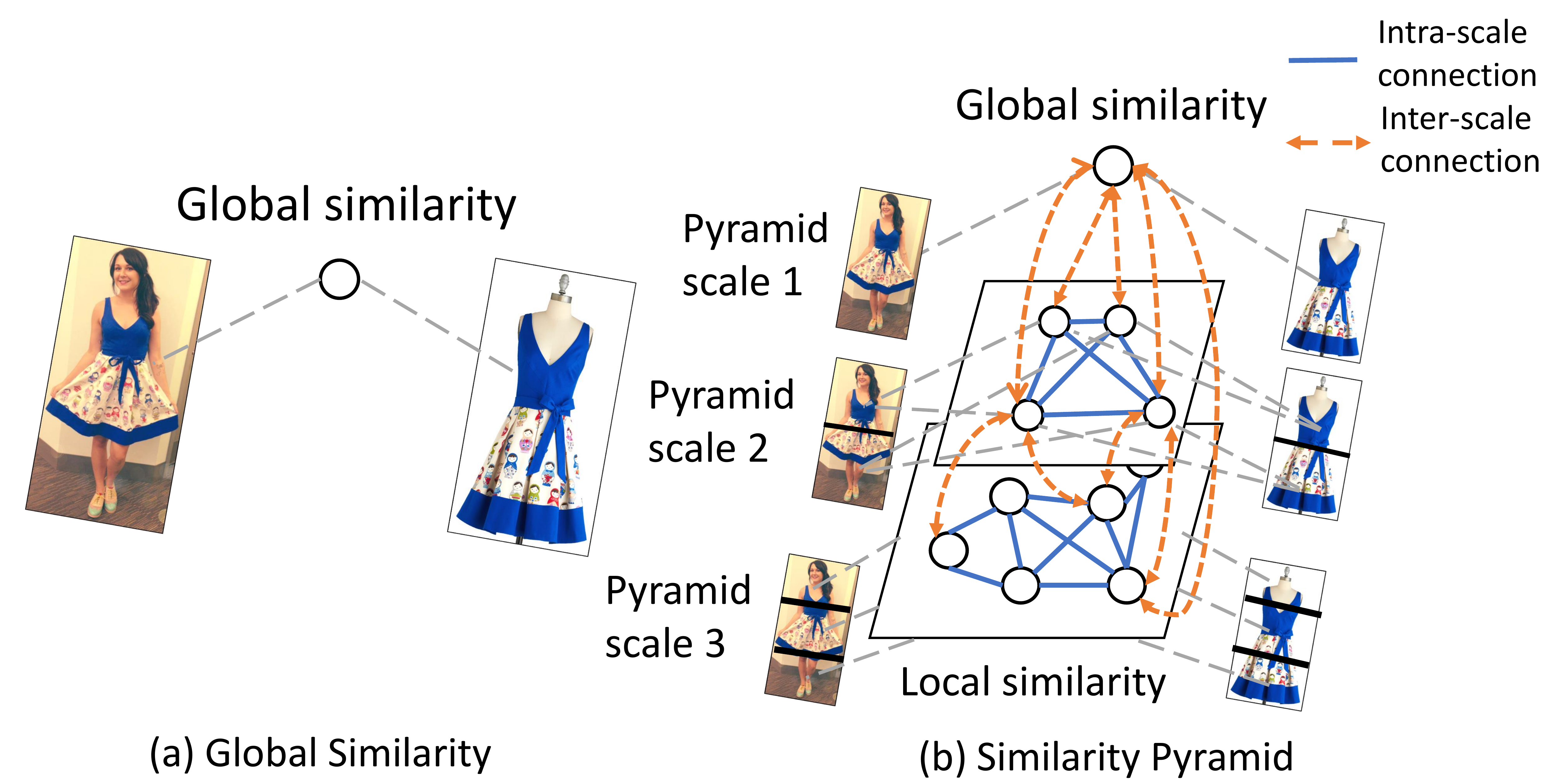}
\label{fig:fig_intro_hs}
\end{minipage}
\caption{Comparison between global similarity and similarity pyramid with graph reasoning. The left illustrates the global similarity. The right shows the similarity pyramid with graph reasoning, where scale 1 computes the global similarity while scale 2 and 3 compute local similarities between all possible combinations of local patches from one image pair. The dash gray lines indicate one similarity is related to two patches. Pyramid similarities (including the global and the local) are reasoned mutually. The blue lines indicate interactions between similarities at one scale while the red dash lines indicate those across scales (best viewed in color). }
\label{fig:fig_intro}
\end{figure}

Fashion image retrieval between customers and online shopping stores has various applications for E-commerce. Given a street-snapshot of clothing image, this task is to search the same garment item in the online store. It is a key step for future applications such as generating descriptions of clothes, brands, materials, and styles. While matching clothes across modalities appears effortless for human vision, it is extremely challenging for machine vision. The same cloth may exhibit large variations due to occlusions, cropping, and viewpoints. More importantly, garments may differ in small local regions such as logos only.


The task of customer-to-shop clothes retrieval has great progresses~\cite{Huang2015,Kiapour2015,Liu2016,Ji2017,Song2017,Corbiere2017,Garcia2017,Cheng2017,Zhang2018} by using convolutional neural networks (CNNs)~\cite{krizhevsky2012imagenet,He2017,girshickICCV15fastrcnn,he2016deep,ren2015faster}.
Existing methods often employed the global similarity pipeline. For example, they first aggregate local features into compact global features, and then compute global similarities between query and gallery images by using cosine or Euclidean distance (see Figure~\ref{fig:fig_intro} (a)).
In the procedure of global feature aggregation, the discriminative local regions of clothes would be submerged in this global representation.
In contrast, human vision verifies whether two clothes are the same by simultaneously comparing the query and the gallery in terms of both global features such as fabric, colors, textures and categories (\eg ``dress'' or ``t-shirt''), as well as local features such as sleeve, collar, and logos. Moreover, human vision only focuses on common parts between the query and the gallery, while ignores those regions only exist in the query (or the gallery) due to occlusions, cropping or viewpoints. We conjecture that for clothing retrieval and verification, comparing clothes in both global and local ways is complementary.


Inspired by the procedure above, we design a novel \emph{Graph Reasoning Network (GRNet)} on a \emph{Similarity Pyramid} to compare a query and a gallery image both globally and locally at different similarity scales.
As illustrated in Figure~\ref{fig:fig_intro} (b), we extract CNN features for all spatial regions at each pyramid scale. An important problem for matching clothes is that the local clothing regions are often mismatched.
In order to solve misalignment between the query and the gallery, we have to enumerate all the region pairs in the same scale to calculate their similarities.
However, as the local regions are not equally important, similarities between aligned regions should be dominated, while those between misaligned pairs should be ignored.


To this end, we construct a pyramid defined by similarities between clothing regions.
This \emph{similarity pyramid} can be formulated as a \emph{graph}, where each node of the graph is the similarity between two corresponding clothing regions in the same scale, while each edge connected two nodes is the normalized similarity of them.
The final similarity (matching score) between a query and a gallery image can be achieved by reasoning on this graph.
GRNet contains a key component of a graph convolutional network (GCN), which performs graph reasoning by propagating messages between nodes.

\begin{figure}
\begin{minipage}{0.9\linewidth}
\centerline{ \includegraphics[width =1\textwidth]{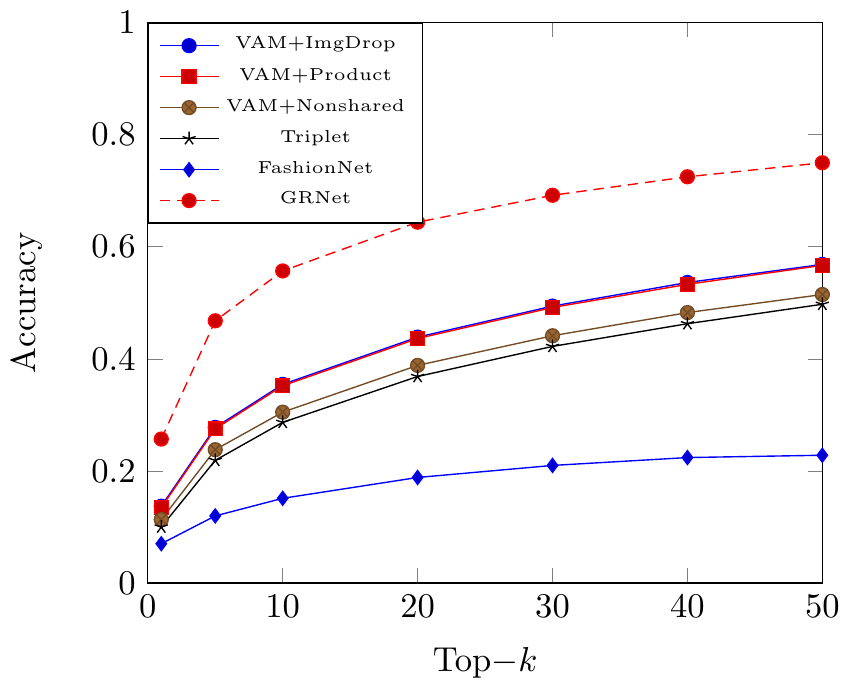}}
\end{minipage}
\caption{Comparison with state-of-the-art methods on DeepFashion consumer-to-shop dataset~\cite{Liu2016}. ImgDrop+GoogleNet and Product+GoogleNet are the best two results ever reported~\cite{Wang2018b}.}
\label{fig:comp_sota}
\end{figure}

The proposed GRNet greatly suppresses the performance degradation caused by occlusions, cropping, viewpoints and small logos, outperforming existing methods with large margins as shown in Figure~\ref{fig:comp_sota}.
Specifically, on the DeepFashion~\cite{Liu2016} benchmark, GRNet \emph{absolutely} improves the top-1, top-20, and top-50 accuracies of the best ever reported results by 12\%, 21\% and 18\%,  and the best results of two state-of-the-art deep matching methods~\cite{Shen_2018_CVPR,Xuan2018}\footnote{We used the codes released by authors and retrained the models on DeepFashion.} by 4\%, 10\%, and 10\% respectively
On Street2Shop~\cite{Kiapour2015} benchmark, GRNet achieves new state-of-the-art results on all five categories \ie ``tops'', ``dresses'', ``skirts'', ``pants'' and ``outerwear''.


Furthermore,
existing benchmarks such as Street2Shop~\cite{Kiapour2015}, DARN~\cite{Huang2015}, and DeepFashion~\cite{Liu2016} have progressed the researches of customer-to-shop clothes retrieval. However, the detailed annotations of occlusions, cropping and views are limited, impeding ablation studies of this task. And they are not suitable to analyze which and how variations affect the retrieval performance.

To this end, we build a new customer-to-shop clothing retrieval benchmark, named \emph{FindFashion}, by revisiting existing datasets, and annotating attributes in terms of occlusions, cropping, and views. FindFashion allows in depth analysis of the impacts of each variation on clothes retrieval. We further introduce four new evaluation protocols of varying difficulties, including \textit{Easy}, \textit{Hard-View}, \textit{Hard-Occlusion}, and \textit{Hard-Cropping}. The splits of training, validation, and test set on FindFashion will be released for fair comparisons.


Our main \textbf{contributions} are summarized in three aspects.
(1) We propose an effective approach for clothing retrieval, \emph{Graph Reasoning Network (GRNet)} on a \emph{Similarity Pyramid}. GRNet computes similarities between a query and a gallery image at different local clothing regions and scales. GRN has an important component of graph convolutional neural network to propagate similarities on the pyramid, performing graph reasoning and producing state-of-the-art performance.
(2) We validate the effectiveness of GRNet on two popular datasets, DeepFashion and Street2Shop. GRNet outperforms state-of-the-art methods with significantly large margins.
(3) We annotate different variations and build a new customer-to-shop retrieval benchmark named FindFashion, which allows the in-depth analysis of the effect of variations for clothing retrieval. Extensive experiments demonstrate that GRNet is more robust against occlusions, cropping, or non-front views than previous methods.


\section{Related Work}

\begin{table}[h]
\scriptsize
\centering
\begin{center}
\begin{tabular}{c|cccc}
\hline
Datasets&Street2Shop~\cite{Kiapour2015}&DARN~\cite{Huang2015}&DeepFashion~\cite{Liu2016}&\textbf{Our} \\ \hline
\#images&416,840&182,780&239,557&565,041 \\
\#pairs&39,479&91,390&195,540&382,230 \\ \hline
Public split&\Checkmark&\ding{55}&\Checkmark&\Checkmark\\
Bbox&\Checkmark\kern-1.5ex\raisebox{1ex}{\rotatebox[origin=c]{125}{\textbf{--}}}&\ding{55}&\Checkmark&\Checkmark\\
View&\ding{55}&\ding{55}&\ding{55}&\Checkmark\\
Occlusion&\ding{55}&\ding{55}&\ding{55}&\Checkmark\\
Cropping&\ding{55}&\ding{55}&\ding{55}&\Checkmark\\ \hline
\end{tabular}
\end{center}
\caption{Comparison of customer-to-shop clothes retrieval datasets.}
\label{tab:table_dataset}
\end{table}

\textbf{Clothes retrieval.} Pioneer work~\cite{XianwangWang2011,Di2013,Fu2012,Garcia2017} on clothing retrieval utilized conventional features such as SIFT and semantic preserving visual phrases. Recently, deep neural networks have been widely applied in clothing retrieval and have pushed the research into a new phase~\cite{Huang2015,Kiapour2015,Liu2016,Ji2017,Song2017,Corbiere2017,Cheng2017,Zhang2018}. These methods usually follow a global similarity computation and matching pipeline, \ie aggregating local features into a single global representation and then performing similarity computation. \cite{Huang2015,Liu2016} explored attributes via multi-task learning to learn representations which are related to specific tags such as ``Crew neck'', ``Short sleeves'' and ``Rectangle-shaped''; \cite{Kiapour2015,Kuo2017} investigated different network architectures which are adept at extracting global features for customer-to-shop clothes retrieval. Instead, \cite{Zhang2018,Corbiere2017} attempted to train models with weakly or noisy supervised signals to reduce the dependency of data annotation and increase the global feature learning efficiency. Recently, \cite{Ji2017} utilized attribute labels to pay more attention to local discriminative regions. Similarly, \cite{Wang2018b} focused on clothes regions and ignored cutter background via a cloth parsing subnetwork. Both the two work employed attention mechanisms in the global feature aggregation to suppress local distractive regions and up-weight the discriminative ones to some extent.  However, they were highly dependent on explicit knowledge such as label and cloth parsing category definition which might be unavailable in real application scenarios. On the contrast, we conduct clothes matching computation via pyramid similarity (including both global and local ones) learning on a relation graph, which can obtain salient component alignment through similarity propagation, and thus achieve more accurate matching. 
Notably, the proposed approach achieves similarities weighting by end-to-end classification training without any explicit supervised signals. Therefore, it is very practical.

There also exist some variants, such as  dialog based clothes search~\cite{Guo2018} , video based clothes retrieval~\cite{Cheng2017}, and attribute feedback based clothes retrieval~\cite{Han2017,Zhao2017}. Their application scenarios and settings are different from ours.

\textbf{Customer-to-shop clothes retrieval datasets.} There exist some customer-to-shop clothes retrieval datasets as listed in Table~\ref{tab:table_dataset}. Kiapour \etal~\cite{Kiapour2015}, collected Street2Shop dataset from a large online retail store. It consists of 78,958 images, 39,479 customer-to-shop pairs, and 396,483 gallery images. Huang \etal~\cite{Huang2015} collected DARN dataset which is composed of upper-clothing images. It has 182,780 images, 91,390 pairs, and 91,390 gallery images, in which only query images are of bounding boxes. However, the training/testing split is not available and thus prevent other research from making a fair comparison.  Liu \etal~\cite{Liu2016} released DeepFashion dataset. It has 239,557 images, 195,540 customer-to-shop pairs, and 45,392 gallery images. It is later revisited for fine grained attribution recognition~\cite{Zakizadeh2018}. All the above datasets are lack of detailed attributes which are most related to clothes retrieval performance. Our benchmark FindFashion contains detailed attribute annotations (\eg views, occlusions and cropping), so that the impacts of attributes on the retrieval performance can be analyzed in detail. 
We have also noticed that there exist other clothes datasets such as~\cite{Bossard2013}, \cite{Chen}, \cite{Zheng2018}, \cite{Inoue2017}, \cite{Lin2018} and \cite{fashionai}. These datasets mainly target at clothes segmentation, attribution prediction and fashion comments, but not customer-to-shop clothes retrieval, and are lack of clothes pairs for evaluation. \cite{Han2017} released Fashion 200k which aims at attribution discovery and clothes retrieval with attribute manipulation,  and is very different from our task.

\textbf{Graph reasoning.} Graph naturally models the dependencies between concepts, which facilitate the research on graph reasoning such as Graph CNN~\cite{Duvenaud2015,Kipf2017,Schlichtkrull2018}, and Gated Graph Neural Network (GGNN)~\cite{Li2016}. These graph neural networks have been widely employed in various tasks of computer vision and have made very promising progress, \eg object parsing~\cite{Liang2017,Liang2016}, multi-label image recognition~\cite{Wang2017a}, visual question answer~\cite{Teney2017}, social relationship understanding~\cite{Wang2018a}, person re-identification~\cite{Shen2018} and action recognition~\cite{Wang}. These work create knowledge graph based on the relationship of different entities, \eg images, objects, proposals, and semantic categories. Instead, we are the first to explore the use of knowledge graph to express the similarity between different pairs of local regions, and apply it to a new field of customer-to-shop clothes retrieval. It can realize the weighting of local region pairs and the enhancement of global matching through the iteration of propagation between pyramid similarities relations, and thus obtain more accurate matching computation. 

\textbf{Image retrieval.} Our work is related to image retrieval approaches~\cite{noh2016large,gordo2016end,Arandjelovic16,radenovic2016cnn,gordo2016deep,tolias2015particular,Yandex2016,chen2017acmmm,chen2019acmmm}.
They target at retrieving rigid objects such as buildings, or scenes,
and often aggregate regional features  into compact representations to compute global similarities.
Different from them, our GRNet aims at retrieving more challenging non-rigid clothes. Moreover, our GRNet captures both local and global similarities, and conducts graph reasoning on a similarity pyramid.

\textbf{Metric learning.} Our work is also related to general deep metric learning~\cite{Opitz2017a, Yuan2017,Kim2018,Opitz2017,Xuan2018}. However, they only conducted experiments on InShop clothes retrieval dataset while our work focuses on customer-to-shop clothes retrieval which is much more challenging as analyzed in~\cite{Liu2016}. We have also compared the proposed GRNet with the state-of-the-art method \cite{Xuan2018} in our experiments.

\section{Methods}

\begin{figure*}[t]
\centering
  \includegraphics[width=0.9\linewidth]{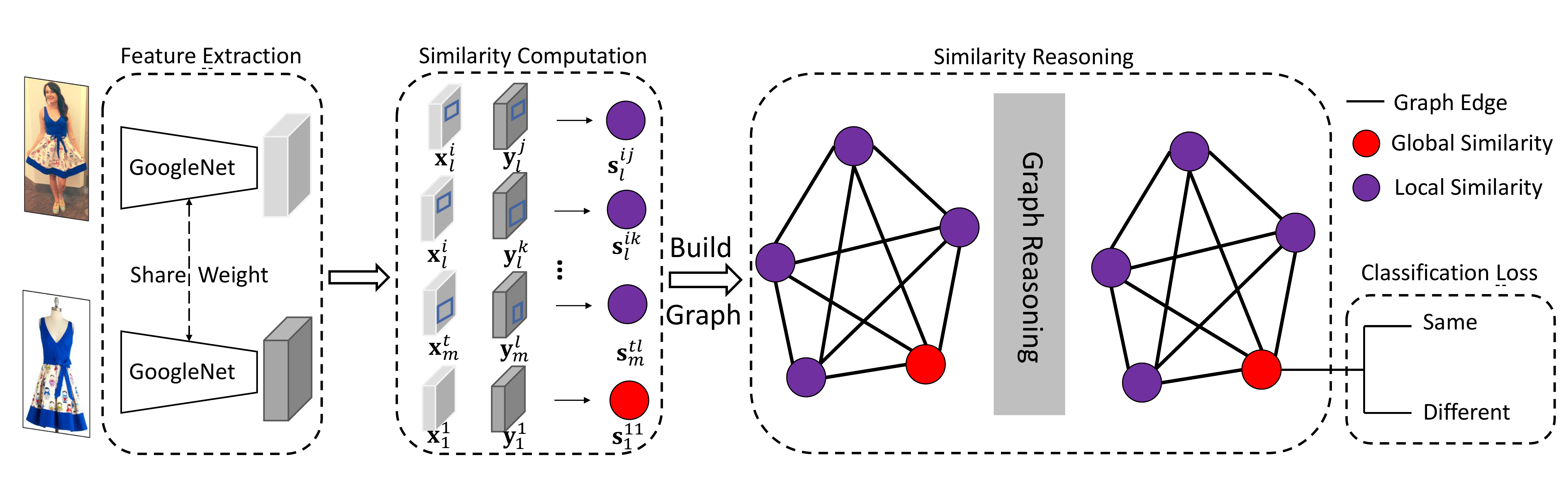}
\caption{The overall framework of the proposed GRNet. Given one query and gallery pair,
their features extracted by deep convolutional networks are fed into Similarity Computation to build a similarity pyramid graph with all region pair similarities being the graph nodes.
In the Similarity Computation, $\mathbf{x}_l^i$ is the $i^{th}$ local feature of the query at scale $l$ while  $\mathbf{y}_l^j$ is the $j^{th}$ one of the gallery, and $\mathbf{s}_l^{ij}$ is their similarity vector.
Further, the global and local similarities are propagated and updated via Similarity Reasoning.
It finally outputs whether the input image pair belong to the same cloth or not.}
\label{fig:framework}
\vspace{-2mm}
\end{figure*}


\subsection{Motivation}
The setup of the customer-to-shop clothes retrieval is the following. Given one customer clothes image query $\mathbf{x}$ and one shop clothes gallery set $\mathbb{G}=\{\mathbf{y}\}$, it computes the similarities $s$ between $\mathbf{x}$ and $\mathbf{y}$, and ranks them. $\mathbf{x}=\{\mathbf{x}^i\}$ and $\mathbf{y}=\{\mathbf{y}^i\}$, where $\mathbf{x}^i\in \mathbb{R}^{C\times 1}$ and $\mathbf{y}^i\in \mathbb{R}^{C\times 1}$ are local features of the customer clothes image and the shop one respectively. Previous customer-to-shop clothes retrieval approaches~\cite{Huang2015,Kiapour2015,Liu2016,Ji2017,Song2017,Corbiere2017,Garcia2017,Cheng2017,Zhang2018}  adopt the following global similarity as:
\begin{equation}
s_g=S_g(A(\mathbf{x}),A(\mathbf{y})),
\label{eq:eq_g}
\end{equation}
where $A(\cdot)$ is the aggregation function and $S_g(\cdot,\cdot)$ is the scalar global similarity function. The aggregation function is usually the average pooling or max-pooling operator. The similarity function often adopts the cosine similarity or Euclidean distance. Ordinarily, the global similarity can reliably estimate the similarity between the query and the gallery. However, the aggregation function might aggregate noisy features such as clutter background, other objects, or unique regions which can only be observed in the query or the gallery when existing occlusions, cropping or different views. This undoubtedly greatly degrades the clothes retrieval performance.

To suppress the above issues, \cite{Wallraven2003,Boughorbel2014} computed the similarity between the query and the gallery by summing up local similarities between local feature pairs with a greedy strategy as follows:
\begin{equation}
s_l=\sum\limits_{i,j}{w_{l}^{ij}S_l(\mathbf{x}^i,\mathbf{y}^j)},
\label{eq:eq_ls}
\end{equation}
where $S_l(\cdot,\cdot)$ is the scalar local similarity function, and $w_{l}^{ij}$ is the scalar weight of local similarities $S_l(\mathbf{x}^i,\mathbf{y}^j)$, which is given by
\begin{equation}
 w_{l}^{ij}=\begin{cases}
    1, & \text{if $j=\argmaxB_k(S_l(\mathbf{x}^i,\mathbf{y}^k))$}.\\
    0, & \text{otherwise}.
  \end{cases}
\end{equation}
 However, greedily finding local feature pairs easily leads to misalignment, which accumulates errors in the final estimated similarity $s_l$.

We attempt to make full use of both the global and local similarities, and learn the importance of different similarities (\ie $w_l^{ij}$ ) automatically to mitigate the above issues.

\subsection{Graph Reasoning Network}
For each query (or gallery) image, instead of extracting local features $\mathbf{x}^i$ ( or $\mathbf{y}^i$) and global features $A(\mathbf{x})$ (or $A(\mathbf{y})$), we extract multi-scale features at pyramid spatial windows, and obtains $ \{\mathbf{x}_l^i \in \mathbb{R}^{C\times 1}\} $ (or $ \{\mathbf{y}_l^i \in \mathbb{R}^{C\times 1}\} $) with $\mathbf{x}_l^i$ (or $\mathbf{y}_l^i$) being the $i^{th}$ local feature for pyramid scale $l$, where $l \in \{1,\cdots,L\}$ indicates scale index from top to down.  Therefore, $\mathbf{x}_1^{1}$ and $\mathbf{y}_1^{1}$ refer to the global feature vector of the query and that of the gallery (\ie, $A(\mathbf{x})$ and $A(\mathbf{y})$)  respectively.  For each scale $l$, assuming there exist $R_l\times C_l$ local spatial windows for each image, we totally have $\sum_l{R_lC_l}$ pyramid features.

\textbf{Similarity pyramid graph.} We build a  similarity pyramid graph with all region pair similarities being the graph nodes, and the relations between two similarities
being the edges. Formally, given a pair of local feature $\mathbf{x}_l^{i}$ and $\mathbf{y}_l^{j}$ from the same pyramid scale $l$, we compute their similarity vector $\mathbf{s}_l^{ij}\in \mathbb{R}^{D\times 1}$ instead of a similarity scalar in Equation~\ref{eq:eq_g} and ~\ref{eq:eq_ls}, by a vector similarity function given by
\begin{equation}
S_p(\mathbf{x}_l^i,\mathbf{y}_l^j)=\frac{\mathbf{P}\abs{\mathbf{x}_l^i-\mathbf{y}_l^j}^2}{\left\Vert \mathbf{P}\abs{\mathbf{x}_l^i-\mathbf{y}_l^j}^2\right\Vert_2},
\end{equation}
where $\abs{\cdot}^2$ and $\left\Vert{\cdot}\right\Vert_2$ indicate element-wise square and  $l_2-$norm respectively. $\mathbf{P}\in \mathbb{R}^{D\times C}$ is a projection matrix which projects pyramid feature difference vectors from $C$ dimension to a lower $D$ dimension. Similarity vectors are guaranteed to have the same magnitude by performing $l_2-$normalization. For any node pair in the graph $\mathbf{s}_{l_1}^{ij}$ and $\mathbf{s}_{l_2}^{mn}$, we define a scalar edge weight $w_p^{l_1ij,l_2mn}$, which is given by
\begin{equation}
w_p^{l_1ij,l_2mn}=\frac{\exp({(\mathbf{T}_{out}\mathbf{s}_{l_1}^{ij})}^\intercal(\mathbf{T}_{in}\mathbf{s}_{l_2}^{mn}))}{\sum\limits_{l,p,q}{\exp({(\mathbf{T}_{out}\mathbf{s}_{l_1}^{ij})}^\intercal(\mathbf{T}_{in}\mathbf{s}_{l}^{pq}))}},
\end{equation}
where $\mathbf{s}^\intercal$ indicates the transpose of the vector $\mathbf{s}$. $\mathbf{T}_{in}\in \mathbb{R}^{D\times D}$ and $\mathbf{T}_{out}\in \mathbb{R}^{D\times D}$ are the linear transformations of incoming edges and outgoing edges for each graph node respectively. When $l_1=l_2$, $w_p^{l_1ij,l_2mn}$ are intra-scale  edges. \ie, their two connected similarity nodes come from the same scale. When $l_1\ne l_2$, $w_p^{l_1ij,l_2mn}$ are inter-scale  edges. \ie, their two nodes come from different scales. Inter-scale edges enable similarities with different scales to propagate messages from each other. In this way, the similarity pyramid graph is defined as $G=(\mathbb{N},\mathbb{E})$, where $\mathbb{N}=\{\mathbf{s}_l^{ij}\}$ and $\mathbb{E}=\{w_p^{l_1ij,l_2mn}\}$.

\textbf{Similarity reasoning.} We reason the similarity  $\mathbf{s}_l^{ij}$ by conducting  a sequence of similarity propagation, linear transformation, and non-linear activation operator. Concretely, similarity is first propagated as
\begin{align}
\hat{\mathbf{s}}_{l_1}^{ij}&=\sum\limits_{l_2,m,n}{w_p^{l_1ij,l_2mn}\mathbf{s}_{l_2}^{mn}} \label{eq:eq_prop} \\
&=\sum\limits_{l_2,m,n}{w_p^{l_1ij,l_2mn}S_p(\mathbf{x}_{l_2}^m,\mathbf{y}_{l_2}^n)}. \label{eq:eq_prop1}
\end{align}
Then, the linear transformation and the non-linear activation are conducted as
\begin{equation}
\mathbf{h}_{l_1}^{ij}=\text{ReLU}(\mathbf{W}\hat{\mathbf{s}}_{l_1}^{ij}), \label{eq:eq_trans}
\end{equation}
where $\mathbf{W}\in \mathbb{R}^{C^{'}\times D}$ is the learnable parameters. Equation~\ref{eq:eq_prop} and ~\ref{eq:eq_trans} can be easily implemented by graph convolution~\cite{Kipf2017}, followed by the nonlinear ReLU. We iteratively reason the similarity pyramid $T$ times by setting $\mathbf{s}_{l_2}^{mn}$ in the right hand side of Equation~\ref{eq:eq_prop} at current step to $\mathbf{h}_{l_2}^{mn}$ from previous step.

\textbf{End-to-end training.} {We use the cross entropy loss over the final reasoned global similarity vector (\ie, $\mathbf{h}_1^{11}$)  and the ground truth $\bar s$ corresponding to the query and the gallery $(\mathbf{x},\mathbf{y})$ to train the whole network end-to-end. In this way, similarities, and their importance are jointly learned.}

\textbf{Network architecture.} Figure~\ref{fig:framework} illustrates the overall framework of the proposed graph reasoning network. It consists of four modules including feature extraction, similarity computation, similarity reasoning and classification loss. In the feature extraction module, we employ GoogleNet~\cite{Szegedy2014} as the backbone, and extract pyramid features by performing max-pooling on its last convolution activation over spatial windows with different pyramid sizes. Both the query and gallery share the same feature extractor.
In the similarity computation module, we compute the similarity between all possible local feature combinations between the query and the gallery at the same pyramid scale. In the similarity reasoning module, we employ a stack of graph convolution and ReLU operators.

\section{FindFashion}
\begin{table}
\center
\begin{center}
\begin{tabular}{c|c|c|c|c}
\hline
Setups & \textit{E}  &  \textit{HO} & \textit{HC} & \textit{HV} \\ \hline
\#Validation& 125863   & 4920    & 15883      & 47164 \\ \hline
\#Test & 30746   & 1250    & 3883      & 11383 \\ \hline
\end{tabular}
\end{center}
\caption{Statistics of four evaluation setups on FindFashion.}
\label{tab:table_benchmark}
\end{table}

We build a new benchmark named FindFashion by revisiting the publicly available datasets. \ie, Street2Shop~\cite{Kiapour2015}, and DeepFashion~\cite{Liu2016}. We labeled 3 attributes (\ie, occlusions, views, and cropping) which mostly affect clothes retrieval performance.
According to the attributes of query, we divided the benchmark into 4 subsets with different difficulty levels. \ie, \textit{Easy}, \textit{Hard-Cropping}, \textit{Hard-Occlusion}, and \textit{Hard-View}.

We adopt the same evaluation measure, \ie, top-k accuracy, to evaluate the performance as in~\cite{Kiapour2015,Liu2016}.

\textbf{Data Collection and cleaning.}
We first merged the two existing datasets (\ie, Street2Shop~\cite{Kiapour2015}, DeepFashion~\cite{Liu2016}), and formed a large dataset containing 382,230 image pairs and 565,041 images, and then we asked the annotators to screen out the image pairs that are clearly not of the same clothes.

\textbf{Annotations.} Gallery images from Street2Shop have no clothes bounding boxes, we first train a Faster RCNN~\cite{ren2015faster} detector over DeepFashion to detect their bounding boxes, and then manually correct them. We annotate three attributes (\ie, views, occlusions and cropping) for all images.
For views, we labeled each clothes images as front, side, or back.  Clothes with the yaw angle in $[-45\degree,45\degree]$ are labelled as front,  those with yaw angle in  $(45\degree,135\degree)$ or $(-135\degree,-45\degree)$ are labelled as side while $[135\degree,225\degree]$  as back. For occlusions, clothes with more than 30\% occluded by other things such as other clothes, mobile phone or belt are labelled as occluded otherwise as un-occluded. For cropping, clothes with more than 30\% cropped are labelled as cropped otherwise as un-cropped.



Images in FindFahsion are of large variance in terms of views, cropping, and occlusions. 
8\% of images are cropped. 3\% of them are occluded. Front view, side view, and back view account for 74\%, 20\%, and 6\% respectively.

\textbf{Evaluation protocol.}
As done in~\cite{Liu2016}, we report top-k accuracy to evaluate the retrieval performance. It reflects the quality of the results of a search engine as they would be visually inspected by a user. Four evaluation setups of different difficulty levels are defined according to the query attribute while keeping the gallery unchanged in the test set:

(1) \textit{Easy (E)}, queries are captured from the front view without cropping or occlusion.

(2) \textit{Hard-Cropping (HC)}, queries are with cropping.

(3) \textit{Hard-Occlusion (HO)}, queries are occluded.

(4) \textit{Hard-View (HV)}, queries are of non-frontal view. Namely, side or back view.

We do not split training dataset according to the above four evaluation setups as we found using maximum training data can achieve better results in all the setups. The detailed statistics of our evaluation protocols are listed in Table~\ref{tab:table_benchmark}.

\section{Experiments}
\begin{table}
\footnotesize
\center
\begin{center}
\begin{tabular}{c
|c|c|c
}
\hline
Methods &Top-1&Top-20&Top-50  \\ \hline
FashionNet~\cite{Liu2016}&7.0 &18.8 &22.8    \\ \hline
Triplet~\cite{Wang2018b}&10.0 &37.0 &49.9       \\ \hline
VAM+Nonshared~\cite{Wang2018b}&11.3 &38.8 &51.5 \\ \hline
VAM+Product~\cite{Wang2018b}&13.4 &43.6 &56.7 \\ \hline
VAM+ImgDrop~\cite{Wang2018b}&13.7 &43.9 &56.9 \\ \hline
DREML(192,48)~\cite{Xuan2018} &18.6 &51.0 &59.1          \\ \hline
KPM~\cite{Shen_2018_CVPR} &21.3 &54.1 &65.2           \\ \hline
GRNet&\textbf{25.7} &\textbf{64.4} &\textbf{75.0}           \\ \hline
\end{tabular}
\end{center}
\caption{Comparison with state-of-the-art methods on DeepFashion consumer-to-shop benchmark~\cite{Liu2016}.}
\label{tab:com_DeepFashion}
\end{table}

\begin{table}
\footnotesize
\center
\begin{center}
\begin{tabular}{c|c|c|c|c|c}
\hline
Method & Tops  & Dresses & Skirts & Pants & Outerwear \\ \hline
Kiapour~\etal~\cite{Kiapour2015}  &38.1 &37.1 &54.6 &29.2 &21.0 \\ \hline
VAM+ImgDrop~\cite{Wang2018b}& 52.3 & 62.1   & 70.9  & --    & --        \\ \hline
Trip.~\cite{Wang2018b}& 44.9 & 56.0   & 69.0  & --    & --        \\ \hline
Trip.+Partial~\cite{Wang2018b}& 47.0 & 58.3   & 72.3  & --    & --        \\ \hline
GRNet   & \textbf{58.3} & \textbf{64.2}   & \textbf{72.5}  & \textbf{48.5} & \textbf{38.6}     \\ \hline
\end{tabular}
\end{center}
\caption{Comparison with state-of-the-art methods on Street2Shop~\cite{Kiapour2015} in terms of top-20 accuracy.}
\label{tab:table_Street2Shop}
\end{table}

\begin{table*}[!htb]
\footnotesize
\center
\begin{center}
\begin{tabular}{c|c|c|c|c|c|c|c|c|c|c|ccccc}
\hline
\multirow{2}{*}{~~~\#~~~} &\multicolumn{6}{|c|}{Local similarity} &\multirow{2}{*}{Intra-scale connection} & \multirow{2}{*}{Inter-scale connection} &\multicolumn{3}{|c}{Accuracy} \\ \cline{2-7} \cline{10-12}
&$1\times 2$&$2\times 1$&$2\times 2$&$3\times 1$&$1\times 3$&$3\times 3$&&&top$-1$&top$-20$&top$-50$ \\ \hline
1 &-&-&-&-&-&-&-&-&14.06&47.60&60.62  \\ \hline
2 &\checkmark&\checkmark&\checkmark&-&-&-&\checkmark&\checkmark&22.60&62.71&73.25  \\ \hline
3 &-&-&-&\checkmark&\checkmark&\checkmark&\checkmark&\checkmark&23.96&64.48&74.32  \\  \hline
4 &\checkmark&\checkmark&\checkmark&\checkmark&\checkmark&\checkmark&-&-&24.48&63.85&74.17  \\ \hline
5 &\checkmark&\checkmark&\checkmark&\checkmark&\checkmark&\checkmark&-&\checkmark&24.79&64.17&74.27  \\ \hline
6 &\checkmark&\checkmark&\checkmark&\checkmark&\checkmark&\checkmark&\checkmark&- &24.58&63.85&73.44  \\ \hline
7 &\checkmark&\checkmark&\checkmark&\checkmark&\checkmark&\checkmark&\checkmark&\checkmark&\textbf{25.73}&\textbf{64.38}&\textbf{75.00}  \\
\hline
\end{tabular}
\end{center}
\caption{Ablation experiments on DeepFashion~\cite{Liu2016}.}
\label{tab:table_ablation_exp}
\end{table*}

\subsection{Implementation Details}
Our implementation on customer-to-shop clothes retrieval follows the practice in~\cite{Liu2016}. We train our models with PyTorch.
We perform standard data augmentation with random horizontal flipping. All cropped images are resized to
$224\times 224$ before being fed into the networks. Optimization is performed using synchronous SGD with momentum 0.9, and weight decay 0.0005 on servers with 8 GPUs. The initial learning rate is set to 0.01 and decreased by a factor of 10 every 20 epochs. All compared models including ours are trained using the same training set for 60 epochs. The feature extractor is initialized with its pre-trained model on ImageNet
while the similarity computation module and the similarity reasoning module are randomly initialized as with~\cite{He2015b}.

  In the feature extraction module, we have totally 7 scales including the global one (\ie, $L=7$). The whole spatial window of images is divided into $1\times 1$, $1\times 2$, $2\times 1$, $2\times 2$, $1\times 3$, $3\times 1$ and $3\times 3$ from scale 1 to 7 respectively. In the similarity reasoning module, we use three (\ie, $T=3$) graph convolution layers with channel number $C^{'}$ set to 128. The projection dimension (\ie, $D$) is set to $512$.

We set the batch size to 64 during training. Each batch consists of 32 clothes with 2 images per clothes. The query and gallery pairs of the same clothes construct positive training samples while other combinations negative ones.

\begin{table}
\footnotesize
\center
\begin{center}
\begin{tabular}{c|c|c|c|c}
\hline
\multirow{2}{*}{Projection dim. $D$} &  \multirow{2}{*}{Channel num. $C^{'}$} &\multicolumn{3}{|c}{Accuracy} \\ \cline{3-5}
&&Top-1&Top-20&Top-50 \\ \hline
512&128&25.73&64.38&75.00  \\
512&256&25.52&64.50& 74.43 \\
512&512&25.92&64.75& 75.54 \\ \hline
256&128&24.06&63.02&73.33  \\
256&256&25.10&64.48& 74.17 \\ \hline
128&128&24.69&63.64&74.38
\\
\hline
\end{tabular}
 \vspace{-2mm}
\end{center}
\caption{Impacts of Dimensions.}
\label{tab:table_dim}
\end{table}
\begin{table*}[!htp]
\footnotesize
\center
\begin{center}
\begin{tabular}{c
|c|c|c
|c|c|c
|c|c|c
|c|c|c
}
\hline
\multirow{2}{*}{Methods} & \multicolumn{3}{|c|}{Easy}&\multicolumn{3}{|c|}{Hard-View}&\multicolumn{3}{|c|}{Hard-Occlusion}&\multicolumn{3}{|c}{Hard-Croping} \\ \cline{2-13}
&Top-1&Top-20&Top-50&Top-1&Top-20&Top-50&Top-1&Top-20&Top-50&Top-1&Top-20&Top-50 \\ \hline
Baseline &16.9 &53.6 &67.6 &10.4 &37.8 &53.2   & 4.5  &25.3 &35.8 &7.3 &35.4 &49.9 \\
DREML(192,48)~\cite{Xuan2018}         &20.7 &54.2 &68.2 &17.2 &44.3 &54.0 & 6.3 & 31.3 &43.8 &10.6 &43.4 &55.2 \\
KPM~\cite{Shen_2018_CVPR}         &22.9 &56.2 &69.2 &18.3 &45.8 &55.8 &5.8 &25.5 &35.4 &9.7 &34.8 &46.7 \\
GRNet&\textbf{27.1}&\textbf{65.1}&\textbf{75.2}
& \textbf{23.3}&\textbf{57.9}&\textbf{69.6}&
\textbf{7.8}&\textbf{35.0}&\textbf{45.0}
&\textbf{14.9}&\textbf{48.4}&\textbf{61.1}   \\
\hline
\end{tabular}
\vspace{-2mm}
\end{center}
\caption{Comparison with state-of-the-art methods on FindFashion.}
\label{tab:table_findfashion}
\end{table*}

\subsection{Results on DeepFashion~\cite{Liu2016}}

\begin{figure*}[!t]
\centering
\includegraphics[width=1\linewidth]{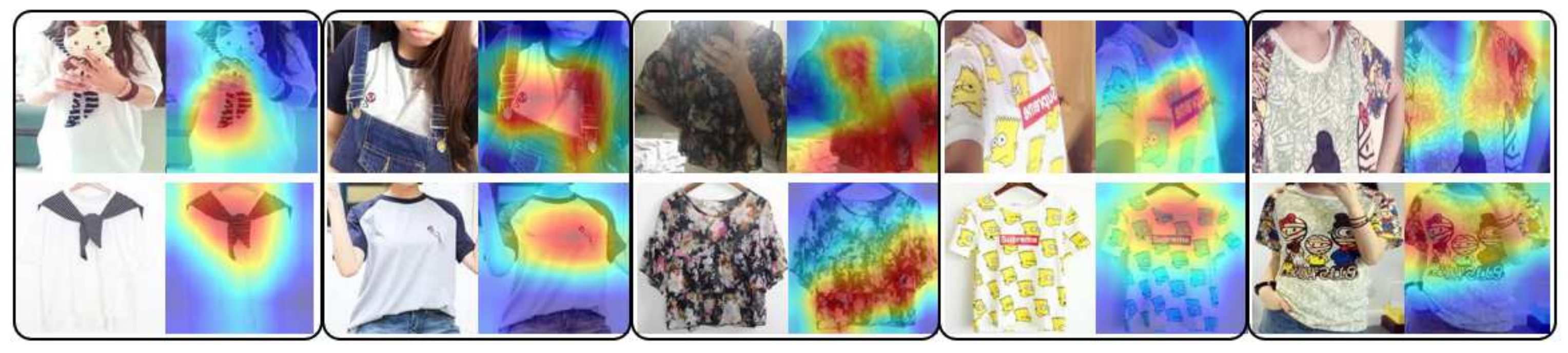}
\caption{Visualization of important regions in the query and the gallery images. Each $2\times 2$ images in one rectangle show one query-gallery image pair and their corresponding highlights, in which the top-left, the top-right, the bottom-left, and the bottom-right are the query, the query highlights, the gallery, and the gallery highlights respectively. Query 1 and 3 are occluded by hands; query 2 is occluded by trousers; query 4 is side view while its gallery front; query 5 is cropped.}
\label{fig:fig_vis_loc}
\vspace{-2mm}
\end{figure*}

\begin{figure}[!b]
\centering
\includegraphics[width=0.5\textwidth]{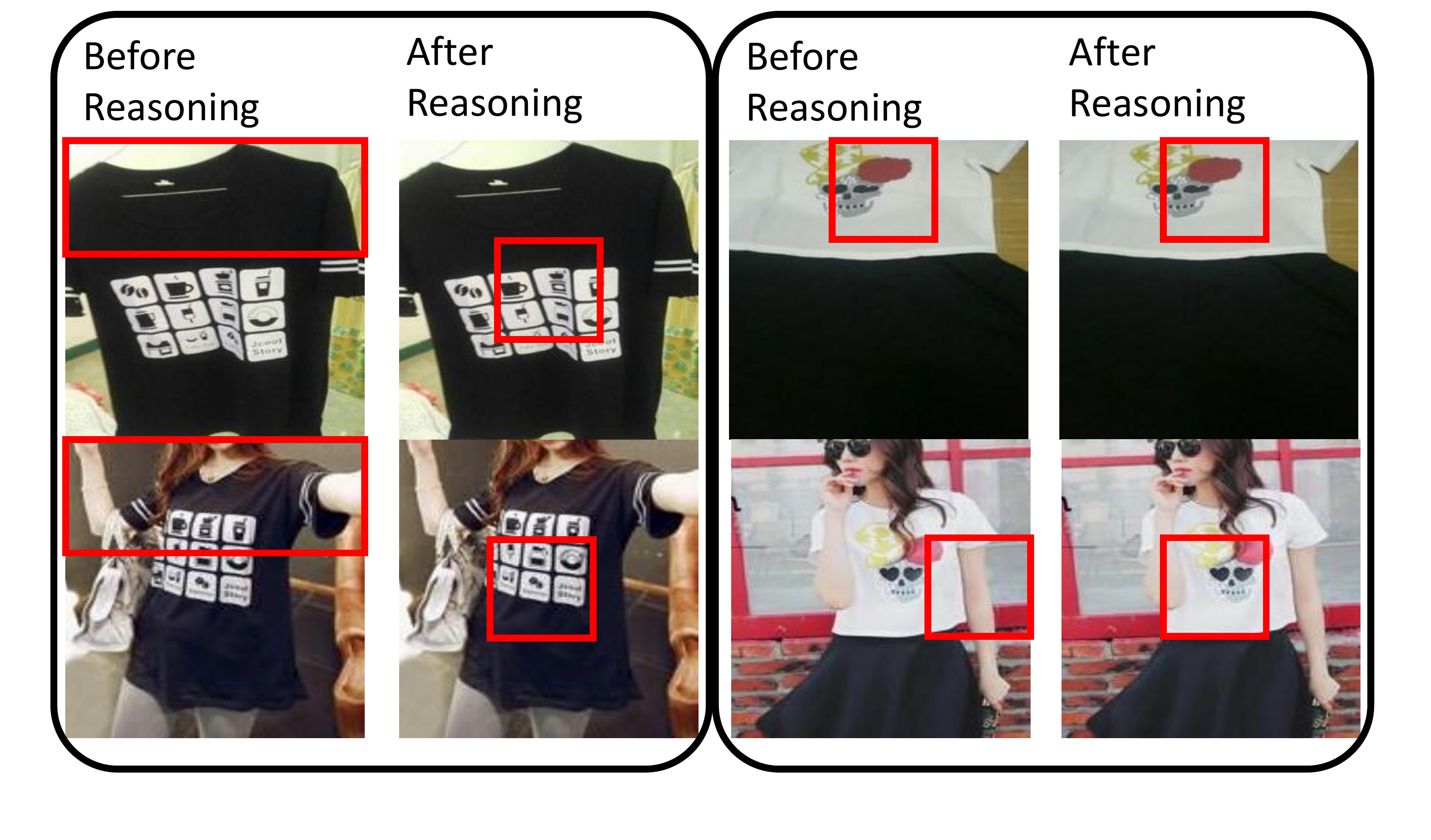}
\caption{Examples of the up-weighted nodes in our similarity pyramid graph. Each node represents one similarity of the local patch (indicated by red rectangles) pair from the query (the top row) and the gallery (the bottom row).     Each $2\times2$ images in one black rectangle show one query-gallery image pair and their up-weighted local patch pairs, where the left column shows the most important node before the similarity reasoning and the right shows it after the similarity propagation. GRNet can up-weight  the similarity between aligned salient clothing components (\eg, logo) after graph reasoning. }
\label{fig:fig_vis_patch}
\vspace{-5mm}
\end{figure}

Table~\ref{tab:com_DeepFashion} compares the proposed GRNet with state-of-the-art methods, including FashionNet~\cite{Liu2016}, triplet-based metric learning approach, and Visual Attention Model (VAM) and its variants (VAM+ImgDrop, VAM+Product, and VAM+Nonshared)~\cite{Wang2018b}, on DeepFashion~\cite{Liu2016}.
Except FashionNet, all counterparts use the same backbone GoogleNet~\cite{Szegedy2014}.
The proposed GRNet outperforms existing methods with an impressive margin. Specifically, it obtains an accuracy of 25.7, 64.4  and 75.0, and absolutely improves the best ever reported results (VAM+Product) by 12\%, 21\% and 18\% respectively. Notably,  VAM uses an attention sub-network which needs clothes segmentation dataset for training. The GRNet is trained with only query-gallery image pairs, thus it is more practical.  We also compare GRNet with DREML~\cite{Xuan2018}, which achieves state-of-the-art performance on multiple general metric learning benchmarks including Inshop~\cite{Liu2016}, recently. We train the DREML model on DeepFashion training set using its open source code with 192 recommended meta classes and 48 ensemble models as done in Table 2 of DREML~\cite{Xuan2018}. Our GRNet is remarkably superior than DREML although DREML employs 48 models for ensemble.
Moreover, we also compare GRNet with KPM~\cite{Shen_2018_CVPR}, which achieves state-of-the-art performance on multiple person re-identification benchmarks and uses the same backbone as our GRNet.
Again, our GRNet outperforms KPM remarkably.
\subsection{Visualization}
To investigate why GRNet works effectively, we employ Grad-CAM~\cite{Selvaraju2017} to visualize the important regions in the query and the gallery images for predicting whether they belong to the same clothes or not in Figure~\ref{fig:fig_vis_loc}. It has been shown that GRNet automatically focuses on local discriminative regions (\eg, scarf, and logo ) and shared regions which can be observed in both the query and the gallery while ignores non-discriminative regions (\eg, non-texture regions), occlusions (\eg, hand) or unique regions which can be observed only in one side due to different views or cropping.
We visualize the similarity node which contributes most to the final classification by selecting the one whose edge outgoing to the global similarity node has the largest weight, in Figure~\ref{fig:fig_vis_patch}. It has been shown that our GRNet can focus on aligned salient clothing components (\eg, logo).


\subsection{Results on Street2Shop}
We compare the proposed GRNet with state-of-the-art customer-to-shop clothes retrieval methods on Street2Shop dataset~\cite{Kiapour2015} in Table~\ref{tab:table_Street2Shop}. It has been shown that it achieves the best results on all five categories of Street2Shop. Particularly, it absolutely improves the best ever reported results by 11.3\% and 5.9\% for tops and dresses categories respectively.

\subsection{Ablation Study}

We investigate the effectiveness of each component in the proposed GRNet by conducting the following ablation studies on DeepFashion dataset~\cite{Liu2016}, shown in Table~\ref{tab:table_ablation_exp}.

\textbf{Graph reasoning.} To validate the effectiveness of graph reasoning, we utilize a GRNet without graph reasoning as our baseline(\#1), which computes the global similarity between global features.
Comparing \#1 and \#7, our graph reasoning acquires $11.6\%$ improvement on the top-1 accuracy.

\textbf{Inter-scale connections}. Comparing \#6 and \#7, it can be observed that the proposed GRNet can achieve $1.15\%$ performance gain on the top-1 accuracy by adding the inter-scale connections (Noted that \#6 and \#4 keep the connections between the global similarity and the local similarities, but removes the connections between different scales).

\textbf{Intra-scale connections}. As reported in Table~\ref{tab:table_ablation_exp}, by propagating similarities at the same scale, our intra-scale connections acquire $0.9\%$ improvement on the top-1 accuracy (\#5 vs \#7).
It shows that the local similarities are also refined by their interactions at the same scale.

\textbf{Multi-scale similarities}. Comparing \#1, \#2, \#3 and \# 7, we observe that the performance is consistently improved when using more scale similarities. Specifically, the accuracy is improved from 14$\%$, 47$\%$ and 60$\%$ to 22$\%$, 62$\%$ and 73$\%$  at top-1, top-20, and top-50 after adding $2\times 1$, $1\times 2$, and $2\times 2$. They are improved slightly by further adding $1\times 3$, $3\times 1$ and $3\times 3$ similarities.
Moreover, we compare the results of different scale levels of local similarity. Comparing \#2 and \#3, the fine scale brings very subtle improvement.
The result shows that the multi-scale similarities can enhance the global similarity representation.


\textbf{Layer number of graph convolution.} We conduct experiments  with different number of graph convolutional layers. The top-1 accuracy increases from $16.8\%$, $22.8\%$, to $25.7\%$ when the number of graph convolutional layer is set to 1, 2, and 3. We observe a performance drop if the layer number is increased further due to over-fitting.
Thus, we fix the graph convolutional layer number to 3.


\textbf{Projection dimension and channel number in graph CNN}. Table~\ref{tab:table_dim} evaluates GRNet with different projection dimension $D$ and channel number $C^{'}$. It has been observed that GRNet is insensitive to projection dimension and channel number. Except
$D=128$, there is no obvious performance drop.  We fix $D=512$ and channel number $C^{'}$ to 128 in all our experiments except otherwise noted.

\subsection{Results on FindFashion}
We evaluate the proposed GRNet on our annotated benchmark FindFashion with four evaluation protocols. Namely, \textit{Easy}, \textit{Hard-View}, \textit{Hard-Cropping}, and \textit{Hard-Occlusion}. We also compare it with DREML~\cite{Xuan2018}, KPM~\cite{Shen_2018_CVPR} and our baseline in Table~\ref{tab:table_findfashion}.
Our GRNet improves the results of the top-20 accuracy up to 65.1 on \textit{Easy}, 57.9 on Hard-View, 35.0 on Hard-Occlusion and 48.4 on Hard-Croping.
Comparing with the results of KPM~\cite{Shen_2018_CVPR} which uses the same backbone as ours, GRNet acquires more improvement on the evaluation protocols of Easy, Hard-View, Hard-Occlusion and  Hard-Croping.
It demonstrates the proposed method's superiority and capability to take full advantages of different scales information to boost the retrieval performance.









\section{Conclusions}
In this paper, we focus on a real-world application task of customer-to-shop clothes retrieval and have proposed a Graph Reasoning Network (GRNet), which first represents the multi-scale regional similarities and their relationships as a graph and then perform graph CNN based reasoning over the graph to adaptively adjust both the local and global similarities. GRNet implicitly achieves alignment and more precise matching of salient clothing components through information propagation among nodes of similarities. To facilitate future research, we have also introduced a new benchmark called FindFashion, which contains rich annotations of clothes including bounding boxes, views, occlusions, and cropping. Extensive experimental results show that our proposed method obtains new state-of-the-art results on both the existing datasets and FindFashion.
\\
\textbf{Acknowledgement} This work was supported in part by Beijing Municipal Science and Technology Commission (Grant No. Z181100008918004), and National Natural Science Foundation of China (Grant No. 61702565).
{\small
\bibliographystyle{ieee_fullname}
\bibliography{egbib}

\begin{thebibliography}{10}\itemsep=-1pt

\bibitem{fashionai}
{Fashionai Dataset. http://fashionai.alibaba.com/datasets}.

\bibitem{Arandjelovic16}
Relja Arandjelovi{\'{c}}, Petr Gronat, Akihiko Torii, Tomas Pajdla, and Josef
  Sivic.
\newblock {NetVLAD: CNN architecture for Weakly Supervised Place Recognition}.
\newblock In {\em CVPR}, 2016.

\bibitem{Bossard2013}
Lukas Bossard, Matthias Dantone, Christian Leistner, Christian Wengert, Till
  Quack, and Luc {Van Gool}.
\newblock {Apparel Classification with Style}.
\newblock In {\em ACCV}, pages 321--335, 2012.

\bibitem{Boughorbel2014}
Sabri Boughorbel and Jean-philippe Tarel.
\newblock {Non-Mercer Kernels for SVM Object Recognition}.
\newblock In {\em BMVC}, 2014.

\bibitem{Chen}
Huizhong Chen, Andrew Gallagher, and Bernd Girod.
\newblock {Describing Clothing by Semantic Attributes}.
\newblock In {\em ECCV}, 2012.

\bibitem{chen2017acmmm}
Zhenfang Chen, Zhanghui Kuang, Kwan-Yee~K. Wong, and Wayne Zhang.
\newblock {Aggregated Deep Feature from Activation Clusters for Particular
  Object Retrieval}.
\newblock In {\em ACM MM Thematic Workshops}, 2017.

\bibitem{chen2019acmmm}
Zhenfang Chen, Zhanghui Kuang, Wayne Zhang, and Kwan-Yee~K. Wong.
\newblock {Learning Local Similarity with Spatial Relations for Object
  Retrieval}.
\newblock In {\em ACM MM}, 2019.

\bibitem{Cheng2017}
Zhi-Qi Cheng, Xiao Wu, Yang Liu, and Xian-Sheng Hua.
\newblock {Video2Shop: Exact Matching Clothes in Videos to Online Shopping
  Images}.
\newblock In {\em CVPR}, pages 4169--4177, 2017.

\bibitem{Corbiere2017}
Charles Corbi{\`{e}}re, Hedi Ben-Younes, Alexandre Ram{\'{e}}, and Charles
  Ollion.
\newblock {Leveraging Weakly Annotated Data for Fashion Image Retrieval and
  Label Prediction}.
\newblock In {\em ICCV Workshop}, 2017.

\bibitem{Di2013}
Wei Di, Catherine Wah, Anurag Bhardwaj, Robinson Piramuthu, and Neel
  Sundaresan.
\newblock {Style Finder: Fine-Grained Clothing Style Detection and Retrieval}.
\newblock In {\em CVPRW}, 2013.

\bibitem{Duvenaud2015}
David Duvenaud, Dougal Maclaurin, Jorge Aguilera-iparraguirre, G Rafael,
  Timothy Hirzel, and Ryan~P Adams.
\newblock {Convolutional Networks on Graphs for Learning Molecular
  Fingerprints}.
\newblock In {\em NIPS}, 2015.

\bibitem{Fu2012}
Jianlong Fu, Jinqiao Wang, Zechao Li, Min Xu, and Hanqing Lu.
\newblock {Efficient Clothing Retrieval with Semantic-preserving Visual
  Phrases}.
\newblock In {\em ACCV}, pages 420--431, 2012.

\bibitem{Garcia2017}
Noa Garcia and George Vogiatzis.
\newblock {Dress Like a Star: Retrieving Fashion Products from Videos}.
\newblock In {\em ICCVW}, pages 2293--2299, 2017.

\bibitem{girshickICCV15fastrcnn}
Ross Girshick.
\newblock {Fast R-CNN}.
\newblock In {\em ICCV}, pages 1440--1448, 2015.

\bibitem{gordo2016deep}
Albert Gordo, Jon Almaz{\'{a}}n, Jerome Revaud, and Diane Larlus.
\newblock {Deep Image Retrieval: Learning Global Representations for Image
  Search}.
\newblock In {\em ECCV}, pages 241--257, 2016.

\bibitem{gordo2016end}
Albert Gordo, Jon Almazan, Jerome Revaud, and Diane Larlus.
\newblock {End-to-end Learning of Deep Visual Representations for Image
  Retrieval}.
\newblock {\em IJCV}, 124(2):237--254, 2017.

\bibitem{Guo2018}
Xiaoxiao Guo, Hui Wu, Yu Cheng, Steven Rennie, and Rogerio~Schmidt Feris.
\newblock {Dialog-based Interactive Image Retrieval}.
\newblock In {\em NIPS}, pages 1--15, 2018.

\bibitem{Han2017}
Xintong Han, Zuxuan Wu, Phoenix~X. Huang, Xiao Zhang, Menglong Zhu, Yuan Li,
  Yang Zhao, and Larry~S. Davis.
\newblock {Automatic Spatially-Aware Fashion Concept Discovery}.
\newblock In {\em ICCV}, pages 1472--1480, 2017.

\bibitem{He2017}
Kaiming He and Ross Girshick.
\newblock {Mask R-CNN}.
\newblock In {\em arXiv preprint arXiv:1703.06870}, 2017.

\bibitem{He2015b}
Kaiming He, Xiangyu Zhang, Shaoqing Ren, and Jian Sun.
\newblock {Delving Deep into Rectifiers : Surpassing Human-Level Performance on
  ImageNet Classification}.
\newblock In {\em ICCV}, 2015.

\bibitem{he2016deep}
Kaiming He, Xiangyu Zhang, Shaoqing Ren, and Jian Sun.
\newblock {Deep Residual Learning for Image Recognition}.
\newblock In {\em CVPR}, pages 770--778, 2016.

\bibitem{Huang2015}
Junshi Huang, Rogerio Feris, Qiang Chen, and Shuicheng Yan.
\newblock {Cross-domain Image Retrieval with a Dual Attribute-aware Ranking
  Network}.
\newblock In {\em ICCV}, pages 1062--1070, 2015.

\bibitem{Inoue2017}
Naoto Inoue, Edgar Simo-Serra, Toshihiko Yamasaki, and Hiroshi Ishikawa.
\newblock {Multi-label Fashion Image Classification with Minimal Human
  Supervision}.
\newblock In {\em ICCVW}, pages 2261--2267, 2017.

\bibitem{Ji2017}
Xin Ji, Wei Wang, Meihui Zhang, and Yang Yang.
\newblock {Cross-Domain Image Retrieval with Attention Modeling}.
\newblock In {\em ACM MM}, pages 1654--1662, 2017.

\bibitem{Kiapour2015}
M.~Hadi Kiapour, Xufeng Han, Svetlana Lazebnik, Alexander~C. Berg, and
  Tamara~L. Berg.
\newblock {Where to Buy It: Matching Street Clothing Photos in Online Shops}.
\newblock In {\em ICCV}, pages 3343--3351, 2015.

\bibitem{Kim2018}
Wonsik Kim, Bhavya Goyal, Kunal Chawla, Jungmin Lee, and Keunjoo Kwon.
\newblock {Attention-based Ensemble for Deep Metric Learning}.
\newblock In {\em CVPR}, 2018.

\bibitem{Kipf2017}
Thomas~N. Kipf and Max Welling.
\newblock {Semi-supervised Classification with Graph Convolutional Networks}.
\newblock In {\em ICLR}, pages 1--14, 2017.

\bibitem{krizhevsky2012imagenet}
Alex Krizhevsky, Ilya Sutskever, and Geoffrey~E Hinton.
\newblock {Imagenet Classification with Deep Convolutional Neural Networks}.
\newblock In {\em NIPS}, pages 1097--1105, 2012.

\bibitem{Kuo2017}
Yin~Hsi Kuo and Winston~H. Hsu.
\newblock {Feature Learning with Rank-Based Candidate Selection for Product
  Search}.
\newblock In {\em ICCVW}, pages 298--307, 2017.

\bibitem{Li2016}
Yujia Li, Richard Zemel, Marc Brockschmidt, and Daniel Tarlow.
\newblock {Gated Graph Sequence Neural Networks}.
\newblock In {\em ICLR}, 2016.

\bibitem{Liang2017}
Xiaodan Liang, Liang Lin, Xiaohui Shen, Jiashi Feng, Shuicheng Yan, and Eric~P
  Xing.
\newblock {Interpretable Structure-Evolving LSTM}.
\newblock In {\em CVPR}, 2017.

\bibitem{Liang2016}
Xiaodan Liang, Xiaohui Shen, Jiashi Feng, Liang Lin, and Shuicheng Yan.
\newblock {Semantic Object Parsing with Graph LSTM}.
\newblock In {\em ECCV}, 2016.

\bibitem{Lin2018}
Wen~Hua Lin, Kuan-Ting Chen, Hung~Yueh Chiang, and Winston Hsu.
\newblock {Netizen-Style Commenting on Fashion Photos: Dataset and Diversity
  Measures}.
\newblock In {\em arXiv preprint}, 2018.

\bibitem{Liu2016}
Ziwei Liu, Ping Luo, Shi Qiu, Xiaogang Wang, and Xiaoou Tang.
\newblock {DeepFashion: Powering Robust Clothes Recognition and Retrieval with
  Rich Annotations}.
\newblock In {\em CVPR}, pages 1096--1104, 2016.

\bibitem{noh2016large}
Hyeonwoo Noh, Andre Araujo, Jack Sim, Tobias Weyand, and Bohyung Han.
\newblock {Large-Scale Image Retrieval with Attentive Deep Local Features}.
\newblock In {\em ICCV}, 2017.

\bibitem{Opitz2017a}
Michael Opitz, Georg Waltner, Horst Possegger, and Horst Bischof.
\newblock {BIER - Boosting Independent Embeddings Robustly}.
\newblock In {\em ICCV}, volume 2017, pages 5199--5208, 2017.

\bibitem{Opitz2017}
Michael Opitz, Georg Waltner, Horst Possegger, and Horst Bischof.
\newblock {BIER: Boosting Independent Embeddings Robustly}.
\newblock In {\em {ICCV}}, 2017.

\bibitem{radenovic2016cnn}
Filip Radenovi{\'{c}}, Giorgos Tolias, and Ondrej Chum.
\newblock {CNN Image Retrieval Learns from BoW: Unsupervised Fine-tuning with
  Hard Examples}.
\newblock In {\em ECCV}, pages 3--20, 2016.

\bibitem{ren2015faster}
Shaoqing Ren, Kaiming He, Ross Girshick, and Jian Sun.
\newblock {Faster r-cnn: Towards Real-time Object Detection with Region
  Proposal Networks}.
\newblock In {\em NIPS}, pages 91--99, 2015.

\bibitem{Schlichtkrull2018}
Michael Schlichtkrull, Thomas~N Kipf, Peter Bloem, Ivan Titov, and Max Welling.
\newblock {Modeling Relational Data with Graph Convolutional Networks}.
\newblock In {\em European Semantic Web Conference}, 2018.

\bibitem{Selvaraju2017}
Ramprasaath~R. Selvaraju, Michael Cogswell, Abhishek Das, Devi {Vedantam,
  Ramakrishna Parikh}, and Dhruv Batra.
\newblock {Visual Explanations from Deep Networks via Gradient-based
  Localization}.
\newblock In {\em ICCV}, 2017.

\bibitem{Shen2018}
Yantao Shen, Hongsheng Li, Shuai Yi, Dapeng Chen, and Xiaogang Wang.
\newblock {Person Re-identification with Deep Similarity-Guided Graph Neural
  Network}.
\newblock In {\em ECCV}, pages 1--20, 2018.

\bibitem{Shen_2018_CVPR}
Yantao Shen, Tong Xiao, Hongsheng Li, Shuai Yi, and Xiaogang Wang.
\newblock End-to-end deep kronecker-product matching for person
  re-identification.
\newblock In {\em CVPR}, June 2018.

\bibitem{Song2017}
Yang Song, Yuan Li, Bo Wu, Chao~Yeh Chen, Xiao Zhang, and Hartwig Adam.
\newblock {Learning Unified Embedding for Apparel Recognition}.
\newblock In {\em ICCVW}, pages 2243--2246, 2017.

\bibitem{Szegedy2014}
Christian Szegedy, Wei Liu, Yangqing Jia, Pierre Sermanet, Scott Reed, Dragomir
  Anguelov, Dumitru Erhan, Vincent Vanhoucke, and Andrew Rabinovich.
\newblock {Going deeper with convolutions}.
\newblock In {\em CVPR}, 2015.

\bibitem{Teney2017}
Damien Teney, Lingqiao Liu, and Anton van~Den Hengel.
\newblock {Graph-Structured Representations for Visual Question Answering}.
\newblock In {\em CVPR}, pages 1--9, 2017.

\bibitem{tolias2015particular}
Giorgos Tolias, Ronan Sicre, and Herv{\'{e}} J{\'{e}}gou.
\newblock {Particular Object Retrieval with Integral Max-pooling of CNN
  Activations}.
\newblock In {\em ICLR}, 2016.

\bibitem{Wallraven2003}
Christian Wallraven and Barbara Caputo.
\newblock {Recognition with Local Features: the Kernel Recipe}.
\newblock In {\em ICCV}, 2003.

\bibitem{Wang}
Xiaolong Wang and Abhinav Gupta.
\newblock {Videos as Space-Time Region Graphs}.
\newblock In {\em ECCV}, 2018.

\bibitem{XianwangWang2011}
Xianwang Wang and Tong Zhang.
\newblock {Clothes Search in Consumer Photos via Color Matching and Attribute
  Learning}.
\newblock In {\em ACM MM}, 2011.

\bibitem{Wang2018a}
Zhouxia Wang, Tianshui Chen, Jimmy Ren, Weihao Yu, Hui Cheng, and Liang Lin.
\newblock {Deep Reasoning with Knowledge Graph for Social Relationship
  Understanding}.
\newblock In {\em IJCAI}, 2018.

\bibitem{Wang2017a}
Zhouxia Wang, Tianshui Chen, Ruijia Xu, and Liang Lin.
\newblock {Multi-label Image Recognition by Recurrently Discovering Attentional
  Regions}.
\newblock In {\em ICCV}, 2017.

\bibitem{Wang2018b}
Zhonghao Wang, Yujun Gu, Ya Zhang, Jun Zhou, and Xiao Gu.
\newblock {Clothing Retrieval with Visual Attention Model}.
\newblock In {\em IEEE Visual Communications and Image Processing}, 2017.

\bibitem{Xuan2018}
Hong Xuan, Richard Souvenir, and Robert Pless.
\newblock {Deep Randomized Ensembles for Metric Learning}.
\newblock In {\em ECCV}, pages 1--12, 2018.

\bibitem{Yandex2016}
Artem~Babenko Yandex and Victor Lempitsky.
\newblock {Aggregating Local Deep Features for Image Retrieval}.
\newblock In {\em ICCV}, 2015.

\bibitem{Yuan2017}
Yuhui Yuan, Kuiyuan Yang, and Chao Zhang.
\newblock {Hard-Aware Deeply Cascaded Embedding}.
\newblock In {\em ICCV}, pages 814--823, 2017.

\bibitem{Zakizadeh2018}
Roshanak Zakizadeh, Michele Sasdelli, Yu Qian, and Eduard Vazquez.
\newblock {Improving the Annotation of DeepFashion Images for Fine-grained
  Attribute Recognition}.
\newblock In {\em arXiv preprint}, 2018.

\bibitem{Zhang2018}
Yanhao Zhang, Pan Pan, Yun Zheng, Kang Zhao, Yingya Zhang, Xiaofeng Ren, and
  Rong Jin.
\newblock {Visual Search at Alibaba}.
\newblock In {\em ACM SIGKDD}, pages 993--1001, 2018.

\bibitem{Zhao2017}
Bo Zhao, Jiashi Feng, Xiao Wu, and Shuicheng Yan.
\newblock {Memory-augmented Attribute Manipulation Networks for Interactive
  Fashion Search}.
\newblock {\em CVPR}, pages 6156--6164, 2017.

\bibitem{Zheng2018}
Shuai Zheng, Fan Yang, M.~Hadi Kiapour, and Robinson Piramuthu.
\newblock {ModaNet: A Large-Scale Street Fashion Dataset with Polygon
  Annotations}.
\newblock In {\em ACM MM}, pages 22--26, 2018.

\end{thebibliography}
}

\end{document}